\documentclass[lettersize,journal]{IEEEtran}  
\usepackage{algorithmic}                                        
\usepackage{array}                                                   
\usepackage{textcomp}
\usepackage{stfloats}
\usepackage{url}
\usepackage{verbatim}
\usepackage{graphicx}
\usepackage{cite}
\usepackage{amsmath,amssymb,amsfonts}
\usepackage[ruled,linesnumbered]{algorithm2e}
\usepackage{algorithmic}
\usepackage{graphicx}
\usepackage{float}
\usepackage{subfig}
\usepackage{textcomp}
\usepackage{booktabs}
\usepackage{multirow}
\usepackage{xcolor}
\usepackage[export]{adjustbox}
\usepackage{tabularx}

\begin{document}
	
	\title{Federated Meta-Learning for Few-Shot Fault Diagnosis with Representation Encoding}
	
	\author{
		Jixuan Cui,~\IEEEmembership{Student Member,~IEEE,} 
		Jun Li,~\IEEEmembership{Senior Member,~IEEE,} 
		Zhen Mei,~\IEEEmembership{Member,~IEEE,} \par
		Kang Wei,~\IEEEmembership{Member,~IEEE,} 
		Sha Wei,~\IEEEmembership{Member,~IEEE,} \par
		Ming Ding,~\IEEEmembership{Senior Member,~IEEE,}  
		Wen Chen,~\IEEEmembership{Senior Member,~IEEE,} 
		Song Guo,~\IEEEmembership{Fellow,~IEEE} 

		\thanks{This work was supported in part by National Key Project under Grant 2020YFB1807700, in part by the National Natural Science Foundation of China under Grant 62071296, in part by the Project on Construction of Intelligent Manufacturing Data Resource public service platform 2021-0174-1-1, in part by Shanghai Fundamental Project under Grant 22JC1404000, Grant 20JC1416502, and Grant PKX2021-D02, in part by the fundings from the Key-Area Research and Development Program of Guangdong Province under Grant 2021B0101400003, in part by Hong Kong RGC Research Impact Fund under Grant R5060-19, and Grant R5034-18, in part by Areas of Excellence Scheme under Grant AoE/E-601/22-R, in part by General Research Fund under Grant 152203/20E, Grant 152244/21E, Grant 152169/22E, and Grant 152228/23E, in part by Shenzhen Science and Technology Innovation Commission under Grant JCYJ20200109142008673. (Corresponding authors: Jun Li, and Zhen Mei.)}

		\thanks{Jixuan Cui, Jun Li, and Zhen Mei are with the School of Electronic and Optical Engineering, Nanjing University of Science and Technology, Nanjing 210094, China (e-mail: jixuancui@njust.edu.cn; jun.li@njust.edu.cn; meizhen@njust.edu.cn).}
		
		\thanks{Kang Wei is with the Department of Computing, The Hong Kong Polytechnic University, Hong Kong 999077, China (e-mail: kangwei@polyu.edu.hk).}
		
		\thanks{Sha Wei is with The Research Institute of Information and Industrialization Integration, China Academy of Information and Communications Technology, Beijing 100191, China (e-mail: weisha@caict.ac.cn).}
		
		\thanks{Ming Ding is with Data61, CSIRO, Sydney, NSW 2015, Australia (e-mail: ming.ding@data61.csiro.au).}
		
		\thanks{Wen Chen is with the Department of Electronics Engineering, Shanghai Jiao Tong University, Shanghai 200240, China (e-mail: wenchen@sjtu.edu.cn).}
		
		\thanks{Song Guo is with the Department of Computer Science and Engineering, Hong Kong University of Science and Technology, Hong Kong 999077, China (e-mail: songguo@cse.ust.hk).}
	}

	\markboth{}
	{Shell \MakeLowercase{\textit{et al.}}: A Sample Article Using IEEEtran.cls for IEEE Journals}
	
	\maketitle
	
	\begin{abstract}
	Deep learning-based fault diagnosis (FD) approaches require a large amount of training data, which are difficult to obtain since they are located across different entities. Federated learning (FL) enables multiple clients to collaboratively train a shared model with data privacy guaranteed. However, the domain discrepancy and data scarcity problems among clients deteriorate the performance of the global FL model. To tackle these issues, we propose a novel framework called representation encoding-based federated meta-learning (REFML) for few-shot FD. First, a novel training strategy based on representation encoding and meta-learning is developed. It harnesses the inherent heterogeneity among training clients, effectively transforming it into an advantage for out-of-distribution generalization on unseen working conditions or equipment types. Additionally, an adaptive interpolation method that calculates the optimal combination of local and global models as the initialization of local training is proposed. This helps to further utilize local information to mitigate the negative effects of domain discrepancy. As a result, high diagnostic accuracy can be achieved on unseen working conditions or equipment types with limited training data. Compared with the state-of-the-art methods, such as FedProx, the proposed REFML framework achieves an increase in accuracy by 2.17\%-6.50\% when tested on unseen working conditions of the same equipment type and 13.44\%-18.33\% when tested on totally unseen equipment types, respectively.
	\end{abstract}
	
	\begin{IEEEkeywords}
		Fault diagnosis, federated meta-learning, security privacy, domain discrepancy, data scarcity.
	\end{IEEEkeywords}
	
	\section{Introduction}
	\IEEEPARstart{F}{ault} diagnosis (FD) plays a significant role in modern industry, ensuring safety and reliability, and preventing breakdowns and losses \cite{9745085}. Traditional FD methods require expert knowledge, such as identifying faults through analyzing abnormal sounds or utilizing signal processing techniques. However, these methods increased the labor intensity and difficulty for equipment operators as the industry continues to grow rapidly in size and frequently change its operational modes. The application of machine learning has made FD intelligent enough to automatically recognize health states and shorten maintenance cycles. Early on, sensitive features are extracted from data and fed into traditional machine learning models, such as expert systems, artificial neural network (ANN), support vector machine (SVM), and others \cite{LEI2020106587}. However, challenges arise from the difficulty of extracting specialized and proper features from large volumes of data and achieving accurate results with the limited learning capability of traditional diagnosis models.\par    
	
	Deep Learning (DL) has become increasingly popular as a solution to address the aforementioned problems as it can learn abstracted representations without human intervention. A novel motor condition monitoring system with 1D convolutional neural network (CNN) was proposed and shown to achieve an elegant classification performance \cite{7501527}. Liu \textit{et al.} proposed a novel method for  FD with recurrent neural network (RNN) in the form of an autoencoder \cite{LIU2018167}. To realize efficient FD with low-quality data, an improved deep fused CNN-based method combined with a complementary ensemble empirical mode decomposition and a short-time Fourier transform was proposed to fully mine fault features \cite{DBLP:journals/ress/ChenLYYG23}. To alleviate the data imbalance in FD, a weakly supervised learning-based method was proposed to introduce real-world samples into the imbalanced dataset \cite{9442330}, and a novel method based on pre-training Wasserstein generative adversarial network with gradient penalty was proposed to generate high-quality faulty samples \cite{CHEN2023112709}. However, these data-driven methods require a large amount of labeled data. Although there is a vast amount of data collected by sensors in modern industry, generating labeled data is very expensive and time-consuming. Moreover, in real industrial scenarios, data is scattered across different entities and privacy-sensitive, rendering it unfeasible to be gathered into centralized servers for training.\par

	Federated learning (FL) \cite{pmlr-v54-mcmahan17a, 10.1145/3298981, 9664296, 9347706,9069945, 9609994, pmlr-v130-haddadpour21a} empowers multiple clients to collaboratively train a global model without compromising data privacy. An FL method for machinery FD with a self-supervised learning scheme was proposed in \cite{zhang2021federated}. The FL framework is also utilized in the context of class-imbalanced FD classification to facilitate the implementation of privacy-preserving functionalities\cite{lu2022class}. However, in practical industrial environments, working conditions and equipment types significantly vary across different companies and change frequently. This means the presence of domain discrepancy extends beyond the traditional demarcation between the training and testing stages. It involves the domain discrepancy between training clients and new clients, with inherent variations observed among individual training clients. As a result, the trained model cannot generalize well to out-of-distribution (OOD) data on new tasks with limited samples, because most learning algorithms heavily rely on the independent and identically distributed assumption on source/target data. Collecting and labeling sufficient data to address this issue is costly and impractical. Data-based approaches such as data sharing and augmentation work well, but may increase the risk of data privacy leakage under the FL framework \cite{ZHU2021371}. \par

	In addressing challenges related to deep model construction and diverse data distributions, an approach with adaptive and independent learning rate design and structure optimization was proposed to enhance both the timeliness of FD and its adaptability to dynamic conditions \cite{9780130}. A novel method combining 2-D-gcForest and $L_{\textnormal{2,p}}$-PCA was proposed to improve the feature representation for different data sources \cite{9760213}. A distribution-invariant deep belief network was proposed to learn distribution-invariant features directly from raw vibration data \cite{8998590}. Transfer learning (TL) is another way against the domain discrepancy problems \cite{LI2022108487}, in which the knowledge from one or more tasks in the source domain can be reused for other related tasks in the target domain. Shao \textit{et al.} developed a fast and accurate FD framework using TL \cite{8432110}. A federated TL framework with discrepancy-based weighted federated averaging was proposed to train a good global FD model collaboratively \cite{9789131}. Meta-learning, a technique that also refers to leveraging previous knowledge to improve learning on new tasks\cite{TIAN2022203}, focuses on learning to learn across a broader range of tasks rather than specific source and target tasks in TL. In industrial scenarios of frequently changing work conditions and equipment types, training a meta-learning model to achieve strong generalization capabilities can greatly meet practical demands. A novel meta-learning method based on model agnostic meta-learning (MAML) \cite{pmlr-v70-finn17a} for FD in rolling bearings under varying working conditions with limited data was proposed in \cite{9749912}. Hu \textit{et al.} proposed a task-sequencing meta-learning method that sorts tasks from easy to difficult to get better knowledge adaptability \cite{9537307}. Moreover, meta-learning could also be combined with semi-supervised learning utilizing unlabeled data for better fault recognition \cite{FENG2022383}.\par

	However, as mentioned earlier, it is challenging to aggregate data from different entities and train models using centralized algorithms above in real-world production environments. In practical and common scenarios, it's necessary to exploit privacy-preserving distributed training algorithms to address the issue of poor diagnosis performance on new tasks caused by domain discrepancy and data scarcity, which has not been fully researched. Furthermore, when considering the inherent domain discrepancies among the data from various participants in FL, it raises a fundamental question: how can we leverage this aspect to strengthen the model's robustness when faced with unobserved tasks?\par
	
	In this study, we tackle this challenge and propose a novel representation encoding-based federated meta-learning (REFML) framework for few-shot FD. REFML harnesses federated meta-learning (FML) and draws inspiration from representation learning for capturing discriminative feature representations \cite{DBLP:journals/corr/abs-1802-07876, NEURIPS2020_24389bfe, 10177379, 9086055, DBLP:conf/ijcai/NozawaS22}. It leverages inherent heterogeneity among training clients by extracting meta-knowledge from different local diagnosis tasks and training a domain-invariant feature extractor in a privacy-preserving manner, effectively transforming it into an advantage for OOD generalization. Without compromising the privacy data of participating clients, the trained model can achieve high performance with very few training samples when encountering new tasks, such as those involving previously unseen working conditions or equipment types, making it well-suited for practical industrial FD scenarios with domain discrepancy and data scarcity problems. \par

	The main contributions of this paper are as follows: \par
		\begin{itemize}    
			\item[1)]
			We propose REFML, an innovative FML-based privacy-preserving method for few-shot FD, a relatively underexplored area in prior research. This approach consists of a novel training strategy based on representation encoding and meta-learning and an adaptive interpolation module.
		\end{itemize}
		\begin{itemize}    
			\item[2)]
			We develop a novel training strategy based on representation encoding and meta-learning to harness the heterogeneity among training clients to improve OOD generalization in FL with limited training samples. With this strategy, the trained model is capable of capturing domain-invariant features and adapting well to unseen tasks to achieve high performance with limited data.
		\end{itemize}
		\begin{itemize}    
			\item[3)]
			We design an adaptive interpolation method by calculating the optimal combination of the local and global models as the initialization of local training. It is capable of mitigating the negative effects of domain discrepancy for better model performance.
		\end{itemize}
		\begin{itemize}    
			\item[4)]
			Experiments are conducted on two bearing datasets and one gearbox dataset. Compared with state-of-the-art methods like FedProx, the proposed REFML framework increases accuracy by 2.17\%-6.50\% when generalizing to unseen working conditions and 13.44\%-18.33\% when generalizing to unseen equipment types, respectively.
		\end{itemize}
	
	The rest of this paper is organized as follows. In section \MakeUppercase{\romannumeral 2} and section \MakeUppercase{\romannumeral 3}, the preliminary work and problem formulation are introduced. Section \MakeUppercase{\romannumeral 4} presents the proposed method in detail. Numerical experiments are conducted in Section \MakeUppercase{\romannumeral 5} to verify the effectiveness of the proposed REFML framework. Section \MakeUppercase{\romannumeral 6} concludes this article.
	~\
	
	\section{ Preliminaries}
	\subsection{Federated Learning}
	FL enables multiple clients to obtain a globally optimized model while safeguarding sensitive data. It generally consists of a central server and multiple clients. The central server manages multiple rounds of federated communication to obtain a global model, extracting valuable information from distributed clients without accessing their private data. Throughout this process, the only elements transmitted are the model parameters. Currently, the prevailing paradigm considers supervised horizontal FL as an empirical risk minimization problem, where the goal is to minimize the aggregated empirical loss, shown as
	\begin{equation}
		\min_{W}  \sum_{u=1}^{U}  \frac{|D_u|}{n} L_{D_u}(W),
		\tag{1}
		\label{eq1}
	\end{equation}
	where $W$ represents the parameters of the global model, and $U$ is the number of total clients. $D_u$ and $|D_u|$ are the local dataset of client $u$ and its size, respectively. The total number of samples is $n=\sum_{u=1}^{U}|D_u|$. $L_{D_u}(W)$ is the empirical loss of client $u$ in the form of an expected risk on local dataset $D_u$ to reflect the model performance, given by
	\begin{equation}
		L_{D_u}(W)=\frac{1}{|D_u|} \sum_{(x, y) \in D_u} l(W(x);y),  
		\tag{2}
		\label{eq2}
	\end{equation}
	where $l(W(x);y)$ is a loss function that penalizes the distance of model output $W(x)$ from label $y$.
	
	\subsection{Meta-Learning}
	Meta-learning, also known as learning to learn, is a technique aiming at enhancing performance on new tasks by utilizing prior knowledge from known tasks. In traditional machine learning, the objective is to train a high-performing model on specific tasks with a fixed algorithm, including artificially designated network architecture, initialization parameters, update methods, and so on. However, in meta-learning, the aim is to find a high-performing algorithm that can adapt well to a set of tasks, particularly those that are unseen.\par
	
	Among different meta-learning methods, MAML is a representative algorithm that focuses on the optimization of initialization parameters. The algorithm trains the model among a series of tasks $\{\varGamma_i\}_{i=1}^N$  from the distribution of $p(\varGamma)$ to obtain high-quality initialization parameters, aiming to perform well on new tasks after training on a few labeled samples. Each task $\varGamma _i$ $(i=1,2,...,N)$ comprises training and testing samples, known as support set $D_{\varGamma _i}^s$ and query set $D_{\varGamma _i}^q$, respectively. \par
	
	Firstly, MAML makes a fast adaptation from $W$  to $W'_i$ in each task's support set $D_{\varGamma _i}^s$ , which can be represented by
	\begin{equation}
		W'_i = W - \alpha \bigtriangledown_W L_{D_{\varGamma _i}^s} (W),
		\tag{3}
		\label{eq3}
	\end{equation}
	where $\alpha$ is the learning rate of this fast adaptation. $L_{D_{\varGamma _i}^s} (W)$ is computed in the form of an expected risk on $D_{\varGamma _i}^s$, given by
	\begin{equation}
		L_{D_{\varGamma _i}^s} (W) = \frac{1}{|D_{\varGamma _i}^s|} \sum_{(x, y) \in D_{\varGamma _i}^s} l(W(x);y),
		\tag{4}
		\label{eq4}
	\end{equation}
	where $|D_{\varGamma _i}^s|$ is the size of the support set, $l$ is the cross entropy loss function. \par
	
\begin{figure}[h]
	\centering
	\vspace{0.3cm}\includegraphics[width=0.6\columnwidth]{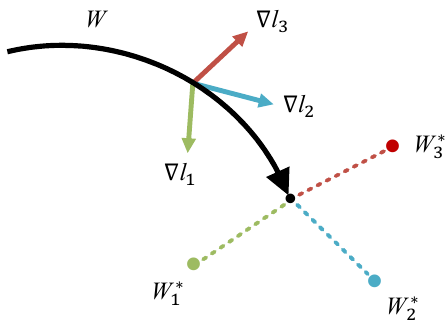}
	\caption{The learning process of MAML. The model is trained to learn general knowledge for adapting to similar, especially unseen tasks quickly and effectively.}
	\vspace{0.3cm}
	\label{fig1}
\end{figure}

	Then, the performance of the adapted parameters $W'_i$ on task $\varGamma _i$ is evaluated on its query set $D_{\varGamma _i}^q$ in the form of empirical loss which reflects the generalization ability of $W$. Hence, the optimization objective is
	\begin{equation}
		\min_{W} \sum_{\varGamma _i \sim p(\varGamma)}   L_{D_{\varGamma _i}^q}(W'_i). 
		\tag{5}
		\label{eq5}
	\end{equation}
	The aggregated loss values of each task are used to update the model parameters $W$, given as
	\begin{equation}
		W = W - \beta \bigtriangledown_W\sum_{\varGamma _i \sim p(\varGamma)}  L_{D_{\varGamma _i}^q}(W - \alpha \bigtriangledown_W L_{D_{\varGamma _i}^s} (W)),
		\tag{6}
		\label{eq6}
	\end{equation}
	where $\beta$ is the meta-learning rate. The purpose of training on multiple tasks is to find a high-quality initial model. As shown in Fig. \ref{fig1}, $W$ are the parameters of the model before updating, and $\nabla l_1$, $\nabla l_2$, and $\nabla l_3$ are corresponding update directions of three training tasks. The objective of MAML is not to attain the best possible performance on a single task, which means reaching any one of the three optimal weights $W_1^*$, $W_2^*$, and $W_3^*$ of three training tasks, but rather to converge on parameters that can swiftly adapt to similar, especially unseen tasks.\par
	
	\section{Problem formulation}
	The core problems of FD in practical applications can be summarized as follows: \par
	1)	Security privacy: Data privacy has been widely and highly valued worldwide. Enterprises are increasingly cautious about the application of data, making it difficult to centralize sufficient data from mutually isolated ``data islands" to train a well-performing model. 
	
	2)	Domain discrepancy and data scarcity: Working conditions and equipment types from which data is collected significantly vary across different companies, and change frequently. Domain discrepancy problems not only exist between participating clients and new clients but also between clients participating in FL. Collecting and labeling sufficient data on new tasks is costly and impractical. It is challenging for models to perform well on new tasks with only a small amount of samples.
	
	Under the FL framework, suppose there are $U$ training clients and $V$ testing clients, whose datasets collected under different working conditions or equipment types are regarded as training tasks and testing tasks. It is noteworthy that clients' private data is not allowed to be shared, and each of them has a relatively small amount of data. Let $\{D_u\}_{u=1}^U$, $\{D_v\}_{v=U+1}^{U+V}$ denote the datasets of training clients and testing clients, respectively. The dataset $D_m=\{(x_i,y_i)\}_{i=1}^{|D_m|}$ $(m=1,2,...,U+V)$ of each client $m$ is divided into support $D_m^s=\{(x_i,y_i)\}_{i = 1}^{|D_m^s|}$ and query set $D_m^q=\{(x_i,y_i)\}_{i=|D_m^s|+1}^{|D_m^s|+|D_m^q|}$. Here $|D_m^s| + |D_m^q| = |D_m|$. Vector $x_i \in \mathbb{R}^d$  is a $d$-dimensional real-valued feature vector regarded as the input of the model, and scalar $y_i \in \{1, 2, 3, ..., N\}$ is a class label, where $N$ is the number of categories. 
	The expected loss of the prediction made with the model parameters $W_m$ of client $m$ on its dataset $D_m$ is defined as $L_{D_m} (W_m)$, which could be computed by
	\begin{equation}
		\begin{split}
			L_{D_m} (W_m) &= \frac{1}{|{D_m}|} \sum_{(x, y) \in {D_m}} l(W_m(x);y) \\
			&= - \frac{1}{|{D_m}|} \sum_{(x, y) \in {D_m}}\sum_{c=1}^{N}\textbf{1}_{[y=c]}{\rm log}(W_m(x)), \\
		\end{split}
		\tag{7}
		\label{eq7}
	\end{equation}
	where $l(W_m(x);y)$ is the cross entropy loss function, and \textbf{1} is the indicator function.\par
	In the meta-training phase, the meta-goal is to find parameters $W^*$ that perform well among training clients after fast adaptation, given by
	\begin{equation}
		W^* = \mathop{\arg\min}\limits_{W} \sum_{u=1}^{U}   L_{D_u^q}(W - \alpha \bigtriangledown_W L_{D_u^s} (W)).
		\tag{8}
		\label{eq8}
	\end{equation}
	In the meta-testing phase, the learned parameters $W^*$ are used to initialize models $\{W_v\}_{v=U+1}^{U+V}$ of testing clients. Then the model will use a small number of samples (the support set) to quickly adapt to the task and expect to achieve good diagnostic accuracy on the query set. The optimization objective during the fast adaptation phase can be expressed as 
	\begin{equation}
		\min_{\{W_v\}_{v=U+1}^{U+V}} \sum_{v=U+1}^{U+V}    L_{D_v^s} (W_v).
		\tag{9}
		\label{eq9}
	\end{equation}
	The problem can be formulated as a $N$-way $K$-shot classification task. The term $N$ refers to the number of categories that a meta task needs to classify, while $K$ represents the number of labeled samples in the support set available for each category. Thus, each task has $N \times K$ samples in the support set, which is equivalent to $|D_m^s|=N \times K$ $(m=1,2,...,U+V)$. All the training clients use both the support set and the query set to train their model, and all testing clients use the support set and the query set to fine-tune and test the model, respectively.
	
	\begin{figure*}[t!]
	\centering
	\vspace*{-0.2cm}\includegraphics[width=1.85\columnwidth]{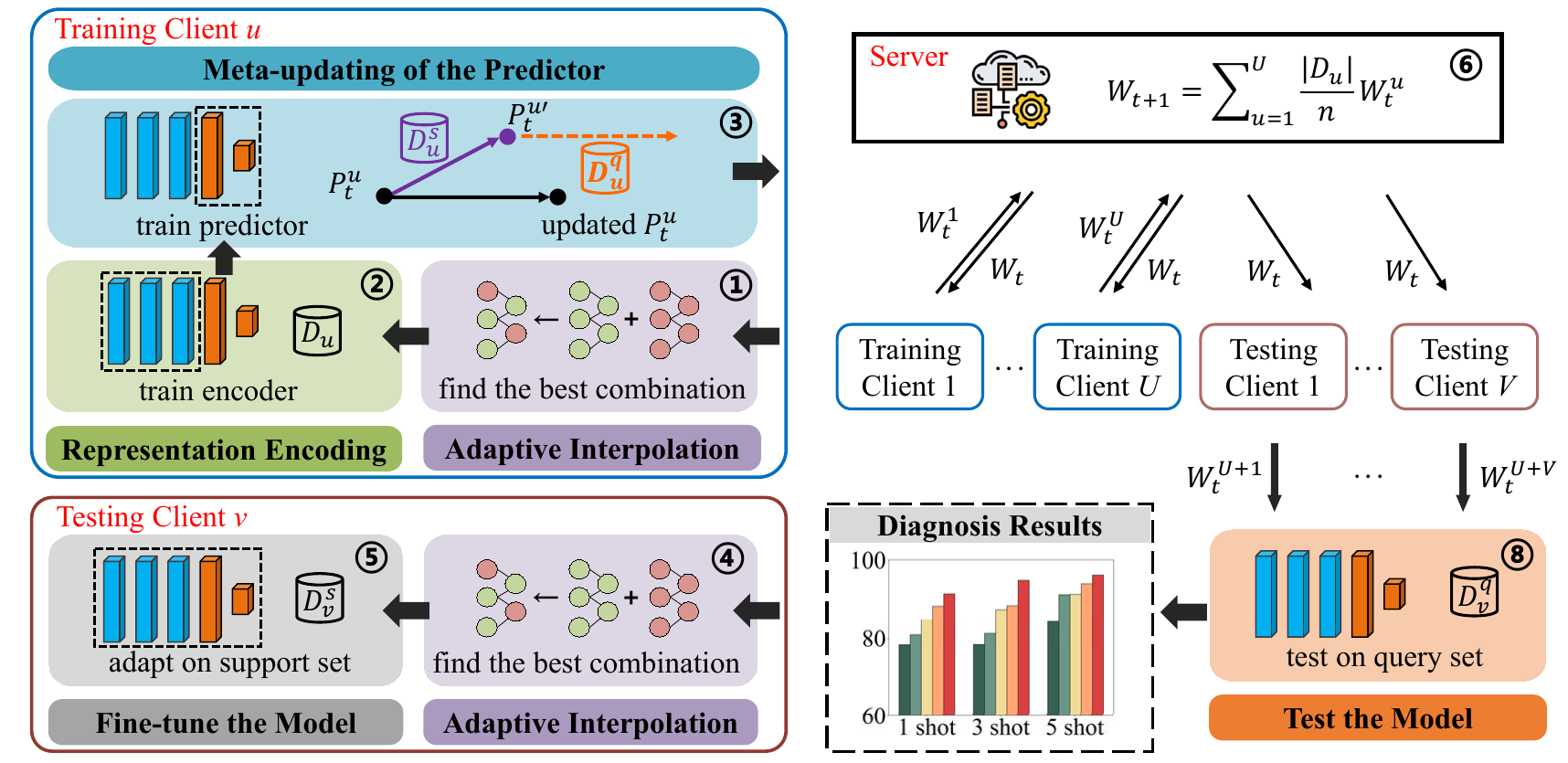}
	\caption{Workflow of the proposed REFML framework. It contains a central server and multiple training and testing clients. The server and clients collaborate through multiple communication rounds to train a model that can effectively adapt to testing clients using extremely limited data with guaranteed privacy protection.}\vspace*{-0.1cm}
	\label{fig2}
\end{figure*}

	\section{Proposed Method}
	This section presents the proposed REFML method for few-shot FD. It consists of two main components: a central server and local clients, which are divided into $U$ training clients and $V$ testing clients. 
	As shown in Fig. \ref{fig2}, the server and clients collaborate through multiple communication rounds to jointly train a model that can adapt well to testing clients with very limited data. In each round, every training client downloads the global model, conducts adaptive interpolation, representation encoding, meta-updating of the predictor, and uploads the model in sequence. Every testing client downloads the global model, conducts adaptive interpolation, and fine-tunes the model. In the following subsections, the proposed training process and the overall workflow of the proposed REFML method will be introduced in detail.\par
	
	\subsection{Adaptive Interpolation}
	In general FL, clients download global model parameters at the beginning of each round and initialize their local models with these parameters. Facing data heterogeneity challenge in FL, a method of mixing global and local models has been proposed to balance generalization with personalization \cite{hanzely2021federated}. Inspired by this, we further exploit personal information in the model communication stage, that is using the optimal interpolation of the global model and the local model as the initial model rather than the global model itself, where the optimal interpolation weights are calculated adaptively using gradient-based search on local data.\par
	
	In communication round $t$, training client $u$ receives global model $W_t$ and aims to find the best combination of the global model $W_t$ and its local model $W_{t-1}^u$ as the initialization of its local training, which is formulated as
	\begin{equation}
		W_t^u = A_t^u \odot W_t  + (O - A_t^u) \odot W_{t-1}^u,
		\tag{10}
		\label{eq10}
	\end{equation}
	where $W_t^u$ is the interpolated model of client $u$ in the $t$-th communication round, and $\odot$ is a Hadamard product. $A_t^u$ are the optimal interpolation weights of the global model $W_t$, whose elements are all between 0 and 1, with the same shape as $W_t$. $O$ is an all-ones matrix, and $O - A_t^u$ are the interpolation weights of the local model. This arrangement ensures that the summation of interpolation weights for the global and local models pertaining to each element position consistently equals 1, thus normalizing the weighting process. \par
	
	To find the optimal interpolation weights $A_t^u$, the interpolation weights $A_{t-1}^u$ of the last round are used to compute a temporary combination $W_t^{u'}$, given by
	\begin{equation}
		W_t^{u'} = A_{t-1}^u \odot  W_t  + (O - A_{t-1}^u) \odot  W_{t-1}^u.
		\tag{11}
		\label{eq11}
	\end{equation}
	Then the temporary combination $W_t^{u'}$ is evaluated on local data, and the interpolation weights can be updated as 
	\begin{equation}
		A_t^u = A_{t-1}^u - \delta \bigtriangledown_{A_{t-1}^u} L_{D_u}(W_t^{u'}),
		\tag{12}
		\label{eq12}
	\end{equation}where $\delta$ is the learning rate.\par 
	Finally, the best combination of the local and global models is computed using (\ref{eq10}) with the updated interpolation weights $A_t^u$. By utilizing this process, every client has the ability to acquire a model that is better tailored to their specific local objective adaptively.\par
	
	\begin{figure}[h]
	\centering
	\vspace*{-0.2cm}\includegraphics[width=0.88\columnwidth]{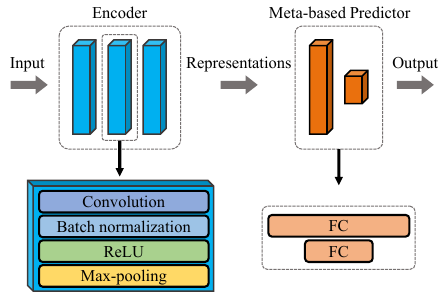}
	\caption{The model structure used in the REFML framework, which consists of an encoder and a meta-based predictor.}
	\label{fig3}
\end{figure}

	\subsection{Representation Encoding}
	As shown in Fig. \ref{fig3}, the related models consist of feature extraction and classification layers, which we refer to as the encoder and the predictor, respectively. We consider the encoder as a domain-invariant representations learner across clients with different data distributions and the predictor as a meta-based classifier learning on a series of training tasks. The encoder contains three convolution units, and each of them is composed of a one-dimensional convolution layer, a batch normalization layer, a rectified linear unit (ReLU) activation layer, and a max-pooling layer. It is responsible for extracting informative feature representations from raw data. \par
	
	Training a specific module within the network while keeping the parameters of other modules unchanged is a common approach to improve the module's suitability for specific tasks, especially when dealing with limited local data. Here, our objective is to enhance the encoder's ability to extract features from local data before meta-knowledge extraction is conducted on the predictor. Once trained, the aggregated encoder can capture domain-invariant features accurately, and the predictor with high-quality initialization parameters can adapt well to new tasks. Correspondingly, the parameters of the model $W_t^u$ are divided into $E_t^u$ and $P_t^u$, representing the parameters of the encoder and predictor, respectively. \par
	
	After adaptive interpolation, each training client makes several local gradient-based updates to find an optimal encoder. For client $u$, it updates encoder $E_t^u$ on local dataset $D_u$ while its predictor $P_t^u$ remains unchanged. The encoder is updated as
	\begin{equation}
		E_{t}^u = E_t^u - \eta \bigtriangledown_{E_{t}^u} L_{D_u}(E_t^u),
		\tag{13}
		\label{eq13}
	\end{equation}
	where $\eta$ is the learning rate. The aggregated encoder will be able to extract common feature abstraction, serving as a shared gripper to obtain domain-invariant features. When encountering unseen working conditions or equipment types, the encoder can extract informative features from previously unseen data types, thereby aiding in achieving better diagnostic performance on new tasks.\par
	
	\subsection{Meta-updating of the Predictor}
	The predictor consists of two fully connected (FC) layers, which manages the classification of the extracted features to generate predictive results for the health states. Following the representation encoding process, we engage in the optimization of the predictor $P_{t}^u$. This optimization is carried out based on the loss computed using parameters that have undergone rapid adaptation, which we refer to as meta-updating. It's important to note that during this meta-updating procedure, the encoder $E_{t}^u$ remains unchanged. First the fast adaptation $P_t^{u'}$ is computed on the support set $D_u^s$, given by
	\begin{equation}
		P_{t}^{u'} = P_t^u - \alpha \bigtriangledown_{P_t^u} L_{D_u^s}(P_t^u).
		\tag{14}
		\label{eq14}
	\end{equation}
	Then the loss of $P_t^{u'}$ on the query set $D_u^q$ is computed to evaluate the performance of meta-parameters $P_t^{u}$ after rapid adaptation. Then the parameters of the predictor will be updated using
	\begin{equation}
		P_{t}^u = P_t^u - \beta \bigtriangledown_{P_t^u} L_{D_u^q}(P_{t}^{u'}),
		\tag{15}
		\label{eq15}
	\end{equation}
	where $\alpha$ is the learning rate of fast adaptation and $\beta$ is the learning rate of meta-updating. Minimizing this loss means that the meta-parameters $P_{t}^u$ will have a better performance after rapid adaptation. In other words, acquiring these adaptation abilities through a series of training tasks constitutes the process of extracting meta-knowledge, which in turn aids in the development of a robust generalization capability. When facing unobserved tasks, informative representations are extracted by the encoder from the raw data, and the predictor with high adaptability utilizes these representations to make accurate diagnoses.\par
	
	\begin{algorithm}[t]
	
	\SetKwData{Left}{left}\SetKwData{This}{this}\SetKwData{Up}{up}
	\SetKwFunction{Union}{Union}\SetKwFunction{FindCompress}{FindCompress}
	\SetKwInOut{Input}{input}\SetKwInOut{Output}{output}
	
	\caption{The proposed REFML method. }
	\label{alg1}
	\Input{Number of communication rounds $T$, number of training and testing clients $U$, $V$, learning rates of the corresponding stages $\delta$, $\eta$, $\alpha$, $\beta$, $\gamma$.}
	\Output{Learned testing models $\{  W^{v}  \}_{v=U+1}^{U+V}$}
	
	Initialize global model $W_0$, local models and interpolation weights of training and testing clients $\{  W_0^{u}  \}_{u=1}^{U}$, $\{  W_0^{v}  \}_{v=U+1}^{U+V}$, $\{  A_0^{u}  \}_{u=1}^{U}$, $\{  A_0^{v}  \}_{v=U+1}^{U+V}$.

	\For{each round $t = 1, 2, ..., T$}{
		
		\For{each training client $u = 1, ..., U$}{
			Compute $A_t^{u}$, $W_t^{u}$ with $D_{u}$ using (\ref{eq10}-\ref{eq12}).\par
			$E_{t}^{u} = E_t^{u} - \eta\bigtriangledown_{E_t^{u}} L_{D_{u}} (E_t^{u})$.\par
		    
		    $P_{t}^{u} = P_t^{u} - \beta \bigtriangledown_{P_t^{u}} L_{D_{u}^q} (P_t^{u} - \alpha \bigtriangledown_{P_t^{u}} L_{D_{u}^s} (P_t^{u}))$.    \par
		    
		}
		\For{each testing client $v = U+1, ..., U+V$}{
			Compute $A_t^{v}$, $W_t^{v}$ with $D_{v}^s$ using (\ref{eq10}-\ref{eq12}).\par
			Fine-tune the model:\par
			$W_{t}^{v} = W_t^{v}- \gamma \bigtriangledown_{W_t^v} L_{D_{v}^s}(W_t^{v})$.  \par
	}
		$W_{t+1} = \sum_{u=1}^{U} \frac{|D_u|}{n} W_{t}^{u}$. 		
	}

\end{algorithm}

	\subsection{The Proposed REFML Algorithm}
	\subsubsection{Local Training}
	The workflow of the proposed REFML framework is depicted in Fig. 2. In each communication round, the training clients engage in a series of operations, including downloading the global model, performing adaptive interpolation, representation encoding, and meta-updating of the predictor, followed by uploading the updated model to the server. On the other hand, the testing clients download the global model, conduct adaptive interpolation, and fine-tune the model using the support set. \par
	\subsubsection{Global Aggregation}
	At the end of each communication round, the server aggregates models uploaded by training clients using
	\begin{equation}
		W_{t+1} = \sum_{u=1}^{U} \frac{|D_u|}{n} W_{t}^{u},
		\tag{16}
		\label{eq16}
	\end{equation}
	where $n$ is the total number of training samples. Subsequently, the server disseminates the aggregated model to all clients in the subsequent round, and all participants collectively iterate through this process until convergence. During the federated communication process, the testing clients have the opportunity to acquire suitable interpolation weights and consistently leverage the domain-invariant feature extraction capability provided by the encoder in the global model, as well as the high-quality initialization parameters of the predictor, to achieve robust diagnostic accuracy on their respective local tasks.\par 
	
	\subsubsection{The Complete Diagnostic Steps}
	The complete process of the proposed REFML method is illustrated in Algorithm \ref{alg1}, and the diagnostic steps are summarized below:\par
	
	1) The training clients download the global model and execute adaptive interpolation;\par
	2) The training clients train the encoder with their local data;\par
	3) The training clients meta-update the predictor and upload the model to the server;\par
	4) The testing clients download the global model and execute adaptive interpolation;\par
	5) The testing clients fine-tune the model with the support set of their local data;\par
	6) The server aggregates the models of all training clients;\par
	7) Repeat steps 1) - 6) until the end of training;\par
	8) The testing clients test the model with the query set of their local data to get diagnosis results.\par
	
	In practice, the computational complexity of our method is comparable to that of typical FL systems, and the training phase does require a certain amount of time. However, it's important to highlight that once the training phase is completed, the inference process becomes highly convenient. This characteristic is particularly advantageous in real-world engineering deployments, where the efficiency of the inference phase frequently surpasses that of the training phase.\par
	
	\begin{table*}[t]
	\centering
	\caption{Working Conditions of Three Datasets.}
	\label{tab1}
	\setlength{\tabcolsep}{3pt}
	\vspace{0.1cm}
	\begin{tabular}{ ccccccccc }
		\toprule
		
		\multirow{2}{*}{Working condition} & \multicolumn{3}{c}{CWRU} & \multicolumn{2}{c}{JNU} &\multicolumn{3}{c}{PHM2009}  \\ 
		& Load(HP) & Speed(rpm) & Size & Speed(rpm) & Size  & Load & Speed(Hz) & Size  \\
		
		\midrule
		
		0 & 0  & 1797 & 2013 &  600  & 488 & High & 30  & 3640 \\
		1 & 1  & 1772 & 2250 &  800  & 488 & High & 35  & 3640 \\
		2 & 2  & 1750 & 2250 & 1000 & 488 & High & 40  & 3640 \\
		3 & 3  & 1730 & 2255 &          &        & High & 45  & 3640 \\
		\bottomrule
		\vspace{0.02cm}
	\end{tabular}
\end{table*}

	\section{Experiments}
	To comprehensively evaluate the effectiveness of the proposed REFML method, two bearing datasets and one gearbox dataset are employed for few-shot scenarios, and the t-distributed stochastic neighbor embedding (t-SNE) technique is applied to compare the feature extraction ability of different methods in the form of visualization. \par
	
	\subsection{Datasets}
	\subsubsection{Case Western Reserve University (CWRU) Dataset}
	The CWRU dataset is a well-known open-source dataset in FD. Its four health states: one normal bearing (NA), inner fault (IF), ball fault (BF), and outer fault (OF), are further classified into ten categories according to three different fault sizes (7, 14, and 21 mils) of each fault state. Each health state corresponds to four distinct working conditions, characterized by varying loads and their corresponding speeds  (1797, 1772, 1750, and 1730 rpm).\par
	
	\subsubsection{JiangNan University (JNU) Dataset}
	The JNU dataset is a bearing dataset acquired by Jiangnan University, China. Four kinds of health states, including NA, IF, BF, and OF, were carried out. Vibration signals were sampled under three rotating speeds (600, 800, and 1000 rpm) corresponding to three working conditions.
	
	\subsubsection{PHM Data Challenge on 2009 (PHM2009) Dataset}
	The PHM2009 dataset is a generic industrial gearbox dataset provided by the PHM data challenge competition. A total of 14 experiments (eight for spur gears and six for helical gears) were performed. It contains five rotating speeds and two loads, corresponding to ten working conditions. Here four experiments of spur gears as four categories are used to conduct the experiments. These experiments are conducted under four different working conditions, with speeds set at 30, 35, 40, and 45 Hz, all operating under high load conditions. The details of the three datasets including corresponding sample sizes are listed in Table \ref{tab1}. However, it is worth noting that in our data scarcity setup, not all the data from the original dataset are utilized. The specific sample size configurations for the experiments can be found in the next subsection.
	
	\begin{table}[t]
	\centering
	\caption{Hyper Parameters.}
	\label{tab2}
	\setlength{\tabcolsep}{3pt}
	\begin{tabular}{ cc }
		
		\toprule
		Hyper parameters      & Value\\
		
		\midrule
		length of sample                           & 1024   \\
		communication round        & 1000   \\
		learning rate                                  & [0.00001, 0.001] \\
		sample each class in support set                            & 1, 3, 5  \\
		sample each class in query set                                 & 10        \\
		
		\bottomrule
	\end{tabular}
\end{table}

	\setcounter{table}{2}
\begin{table}[t]
	\centering
	\caption{Experiment Settings on Unseen Working Conditions.}
	\label{tab3}
	\setlength{\tabcolsep}{3pt}
	\begin{tabular}{ cccc }
		\toprule
		Dataset & Fold & Meta-train conditions & Meta-test condition \\
		\midrule
		\multirow{4}{*}{CWRU} & 1 & 1, 2, 3  & 0  \\
		& 2 & 2, 3, 0  & 1  \\
		& 3 & 3, 0, 1  & 2  \\
		& 4 & 0, 1, 2  & 3  \\
		\multirow{3}{*}{JNU}      & 1 & 1, 2  & 0  \\
		& 2 & 2, 3  & 1  \\
		& 3 & 3, 0  & 2  \\
		\multirow{4}{*}{PHM2009}    & 1 & 1, 2, 3  & 0  \\
		& 2 & 2, 3, 0  & 1  \\
		& 3 & 3, 0, 1  & 2  \\
		& 4 & 0, 1, 2  & 3  \\
		
		\bottomrule
		\vspace{-0.7cm}
	\end{tabular}
\end{table}

\begin{figure} [h]
	\centering
	\subfloat[]{\includegraphics[width=0.50\columnwidth]{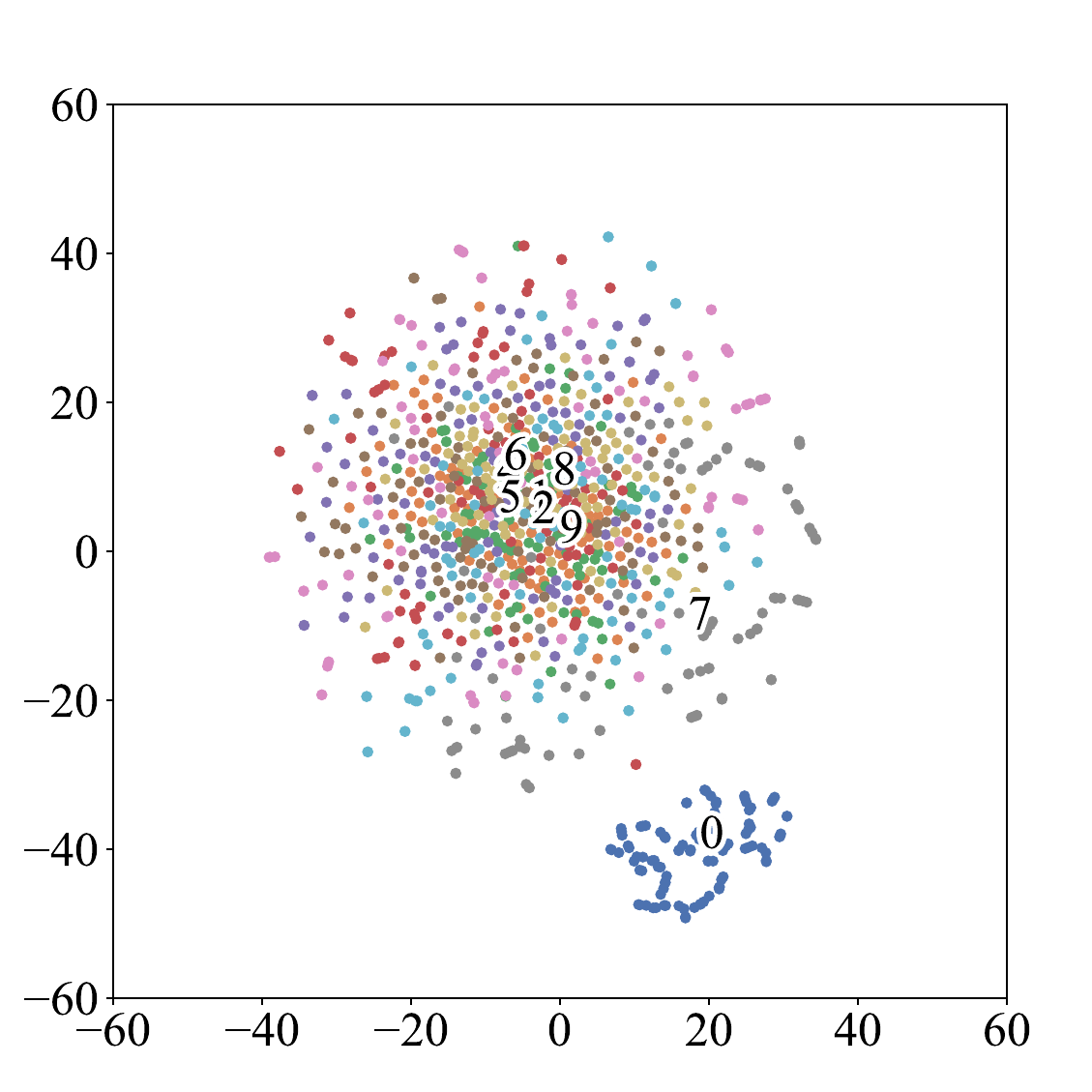}}
	\hspace{-0.25 cm}
	\subfloat[]{\includegraphics[width=0.50\columnwidth]{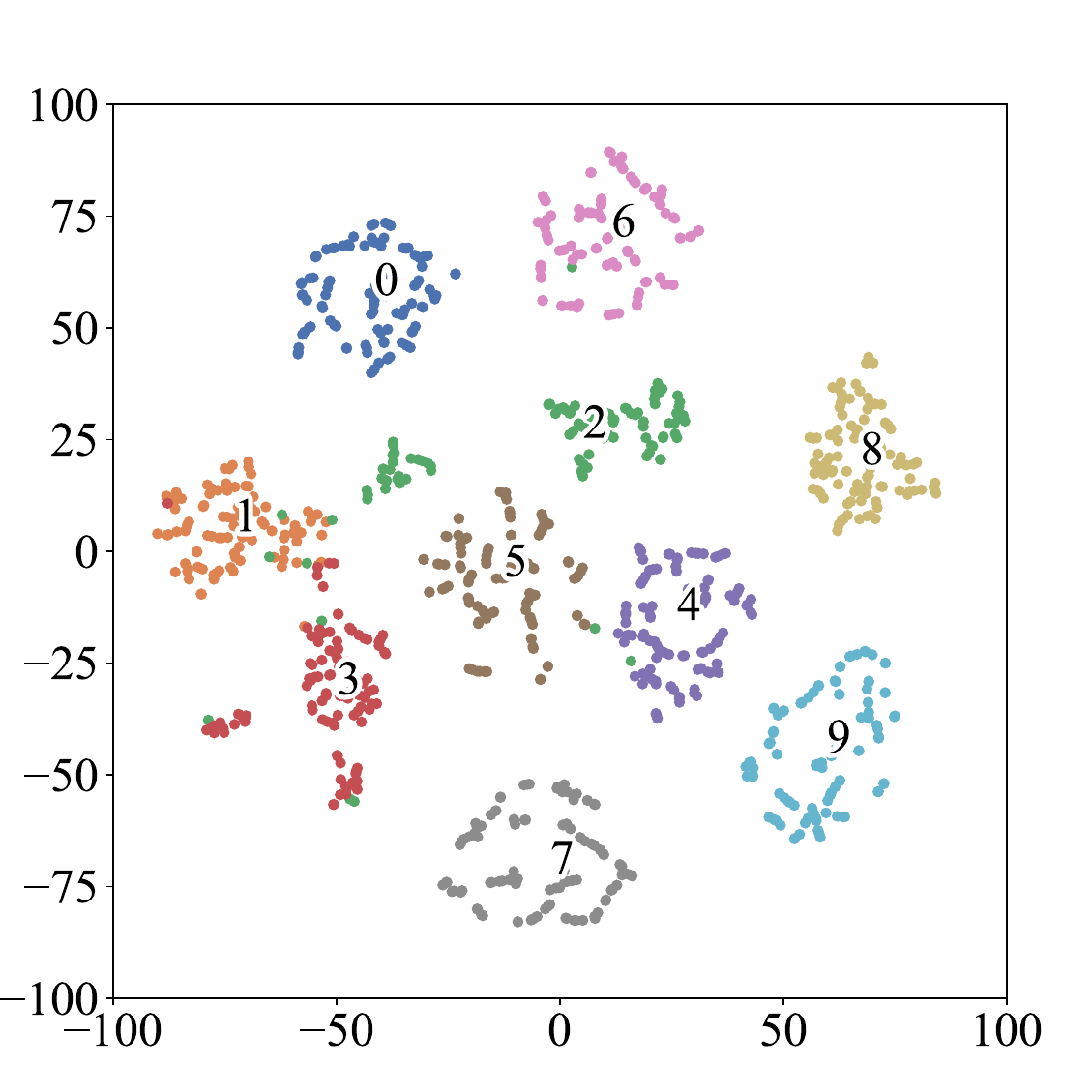}} \\
	\vspace{-0.20 cm} 
	
	\subfloat[]{\includegraphics[width=0.50\columnwidth]{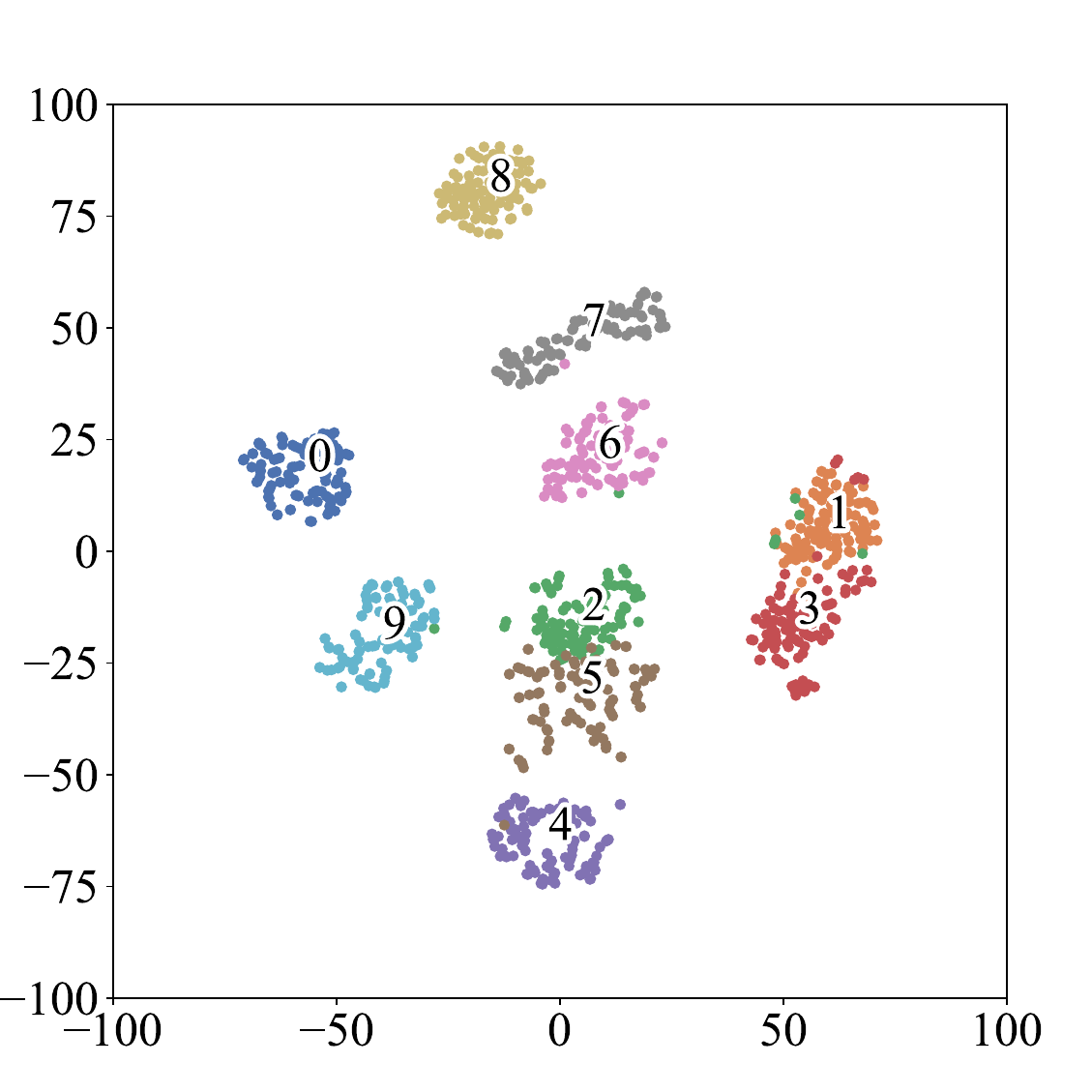}}
	\hspace{-0.25 cm}
	\subfloat[]{\includegraphics[width=0.50\columnwidth]{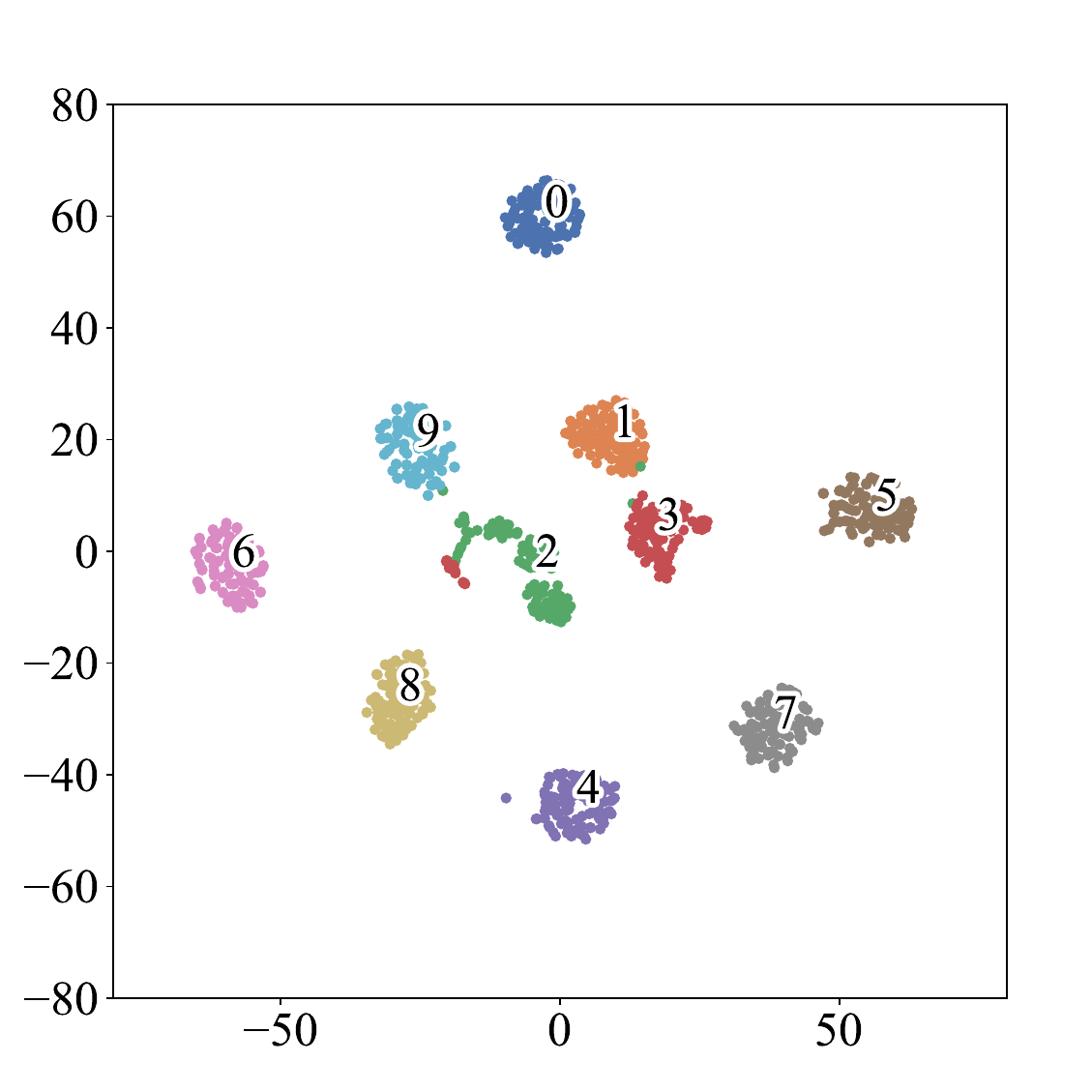}}

	\caption{Visualization of extracted representations using t-SNE. (a) Raw data. (b) FedAvg-FT. (c) FedProx-FT. (d) REFML. }
	\label{fig4} 
\end{figure}

	\begin{figure}[h]
	\centering
	\includegraphics[width=0.82\columnwidth]{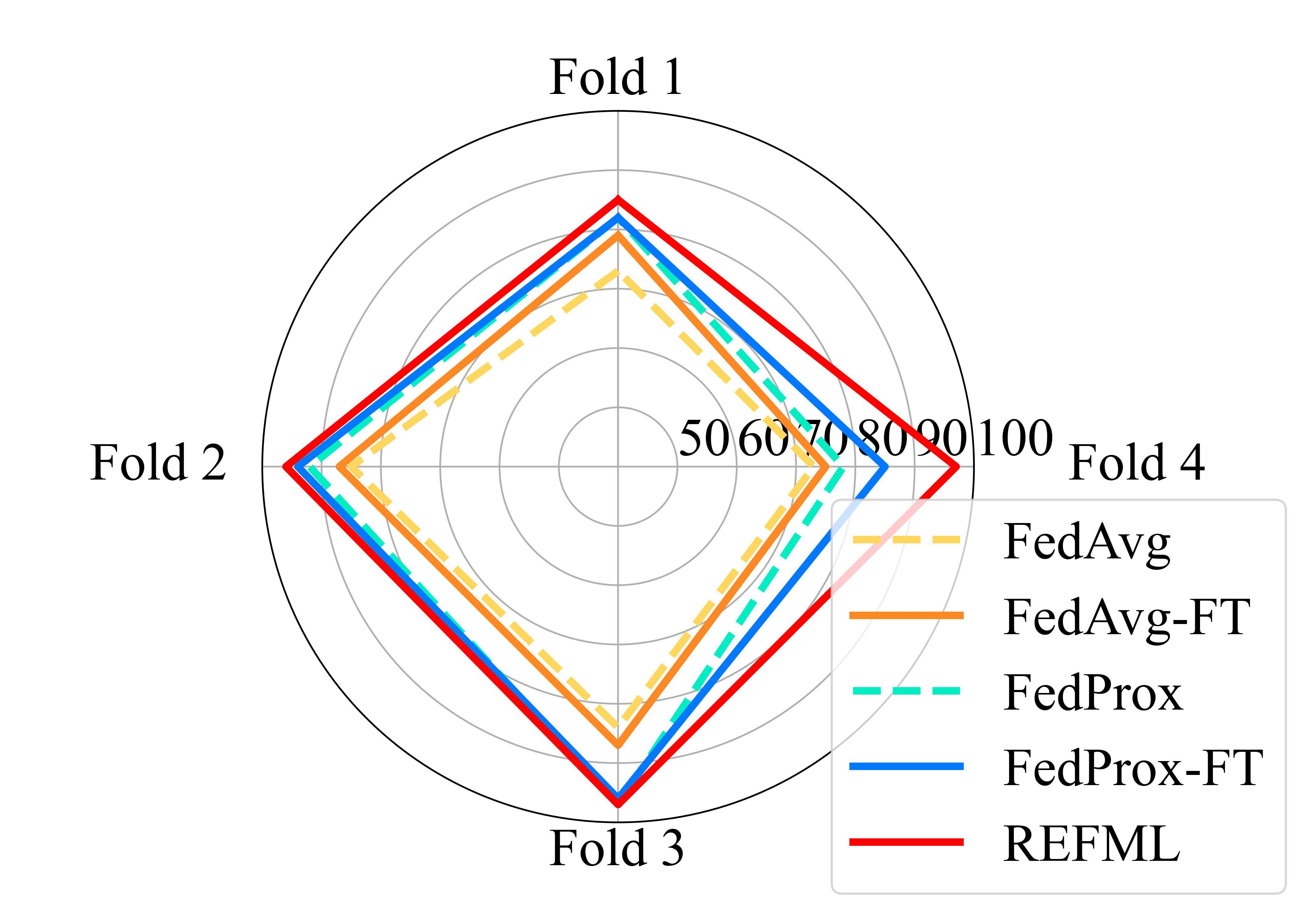}
	\caption{The diagnostic accuracy of the 1-shot sub-experiments on the CWRU dataset.}
	\label{fig5}
	\vspace{-0.1cm}
\end{figure}

	\subsection{Experiment Setup}
	\subsubsection{Network Structure and Hyper parameters}
	In our experiments, related FD models are based on CNN, which contains three convolution units and two FC layers. Each convolution unit is composed of a one-dimensional convolution layer, a batch normalization layer, a ReLU activation layer, and a max-pooling layer. Its output is flattened into a tensor with one dimension and a length of 4096 and then used as input for the following FC layers. Note that the number of the output units $N$ of the last layer varies with the category number of the specific task. 
	\par
	The learning rates are obtained by searching within the range of 0.00001 to 0.001. In the federated communication process, the maximum number of communication rounds is 1000. In the few-shot scenario, the shot number in the query set is 10, and the shot number in the support set varies in the range of 1, 3, and 5. Some crucial hyper parameters are shown in Table \ref{tab2}.
	
	\subsubsection{Baselines}
	We evaluate our approach against several state-of-the-art baselines, including FedAvg\cite{pmlr-v54-mcmahan17a}, FedProx\cite{MLSYS2020_38af8613} and their fine-tuned versions denoted by FedAvg-FT and FedProx-FT for a fair comparison. Fine-tuned versions of these baselines use the support set of the testing clients to fine-tune the model received from the server before testing it on the query set. All of these four baselines use all the data, including support and query set on the training clients.

\setcounter{table}{3}

\begin{table*}[t]
	\centering
	\caption{Experiment Results on Unseen Working Conditions.}
	\label{tab4}
	\setlength{\tabcolsep}{5pt}
	\begin{tabular}{ cccccccccc  }
		\toprule
		\centering\multirow{2}{*}{ Methods} & \multicolumn{3}{c}{CWRU} & \multicolumn{3}{c}{JNU} & \multicolumn{3}{c}{PHM2009}  \\
		& 1-shot & 3-shot & 5-shot & 1-shot & 3-shot & 5-shot & 1-shot & 3-shot & 5-shot \\
		\midrule
		\centering FedAvg                                       & 78.25 & 78.33 & 84.25 & 68.89 & 72.50 & 72.78 & 71.56 & 71.88 & 72.19  \\
		\centering FedAvg-FT                                 & 80.92 & 81.25 & 91.17 & 70.83 & 73.61 & 74.17 & 71.88 & 72.81 & 72.81  \\
		\centering FedProx                                      & 84.75 & 87.25 & 91.25 & 71.67 & 74.17 & 76.67 & 72.50 & 73.44 & 74.38  \\
		\centering FedProx-FT                                & 88.13 & 88.25 & 93.88 & 74.17 & 76.67 & 77.50 & 73.13 & 74.06 & 75.16  \\
		\centering REFML             & \textbf{91.38} & \textbf{94.75} & \textbf{96.05} & \textbf{76.94} & \textbf{80.83}  & \textbf{81.11} & \textbf{75.94} & \textbf{77.19} & \textbf{77.82}\\
		\bottomrule
	\end{tabular}
\end{table*}

\begin{figure*}[t]
	\centering

	\includegraphics[width=1.1\columnwidth]{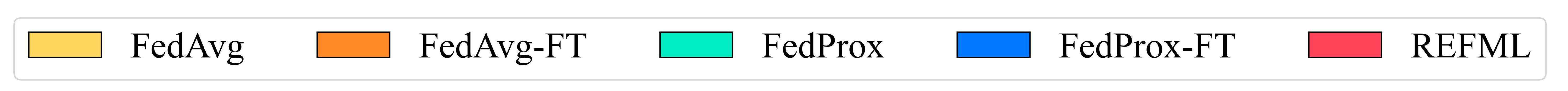} \vspace{-0.3cm} \\
	\subfloat[]{\includegraphics[width=0.67\columnwidth]{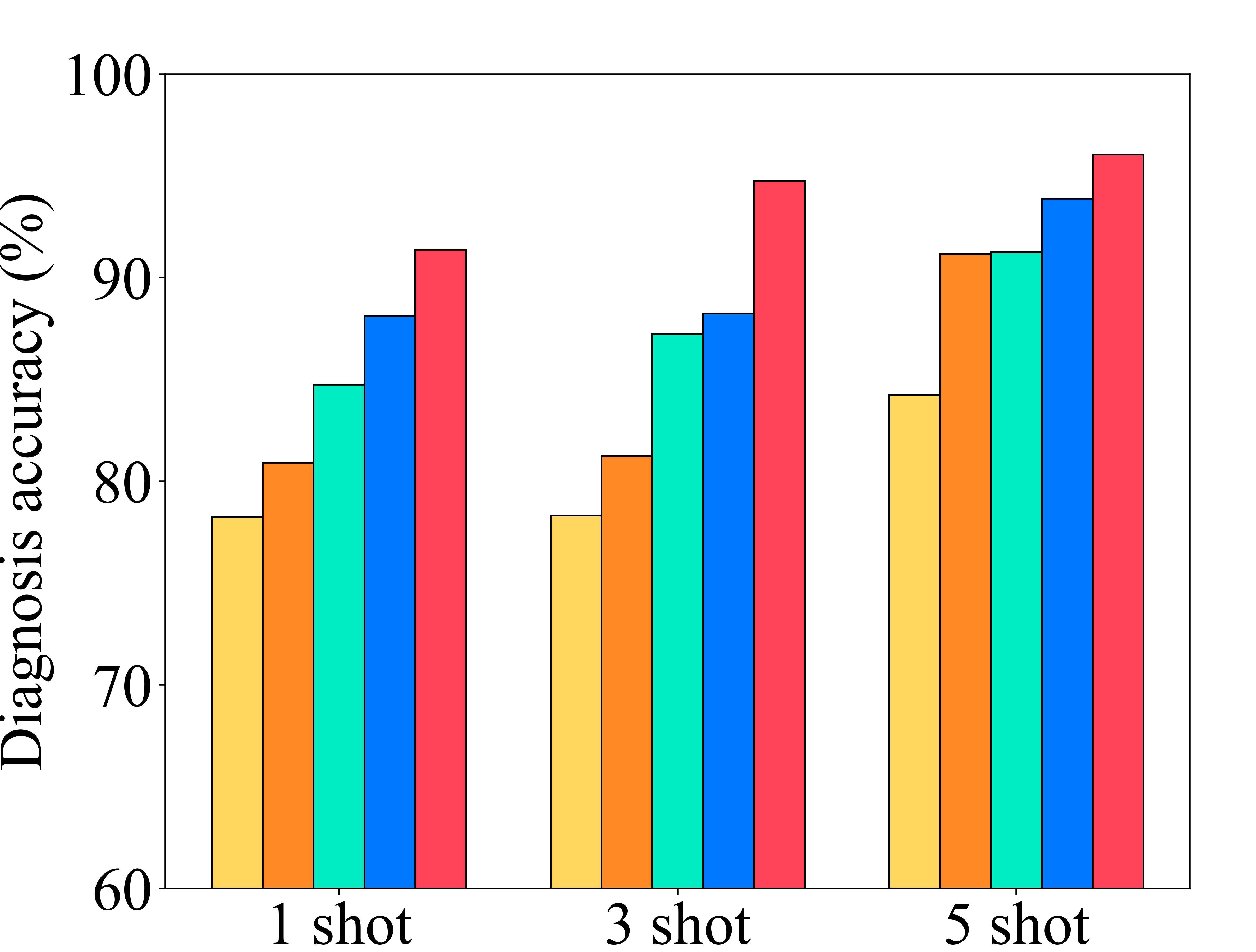}}\vspace{-0.1cm}
	\subfloat[]{\includegraphics[width=0.67\columnwidth]{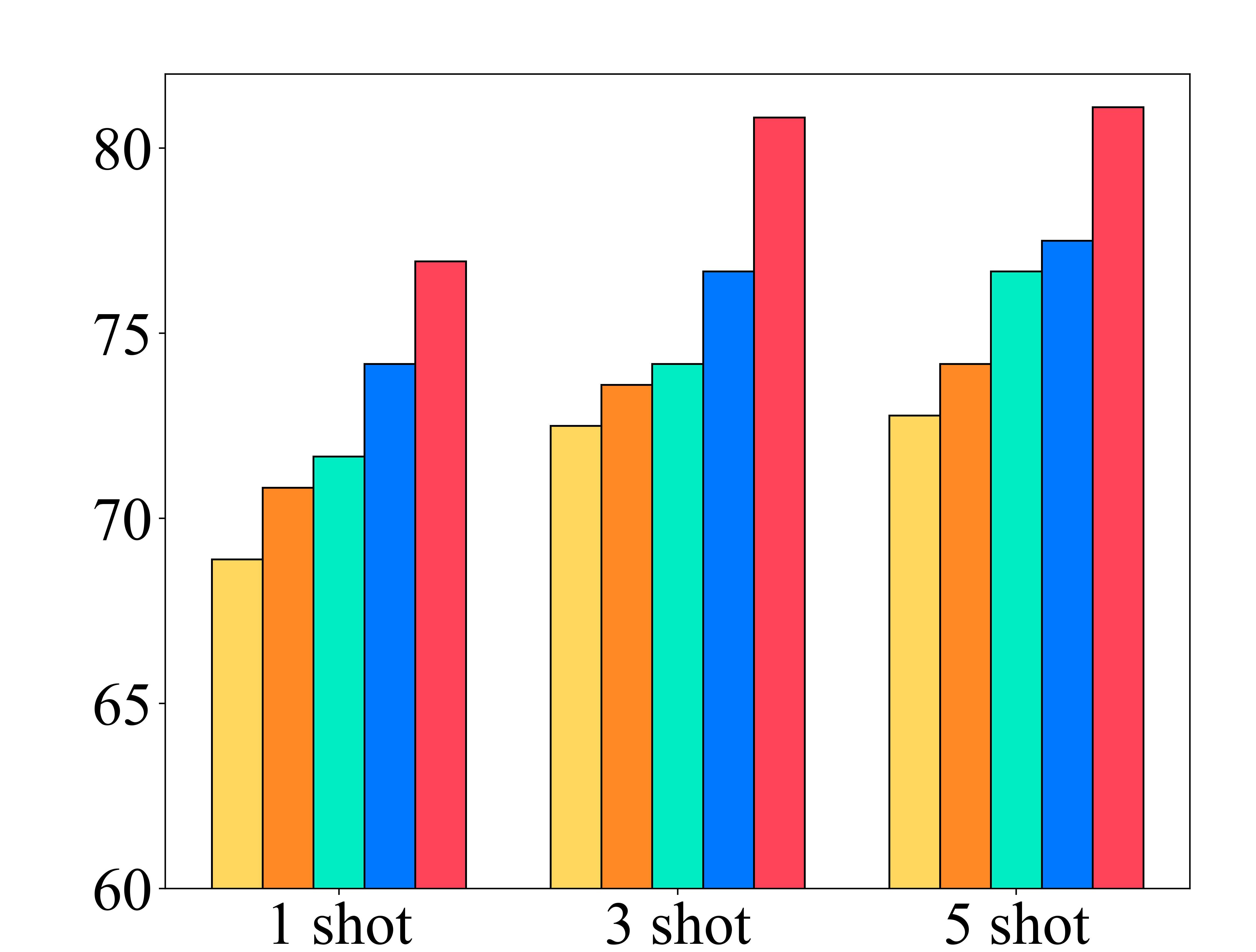} }
	\subfloat[]{\includegraphics[width=0.67\columnwidth]{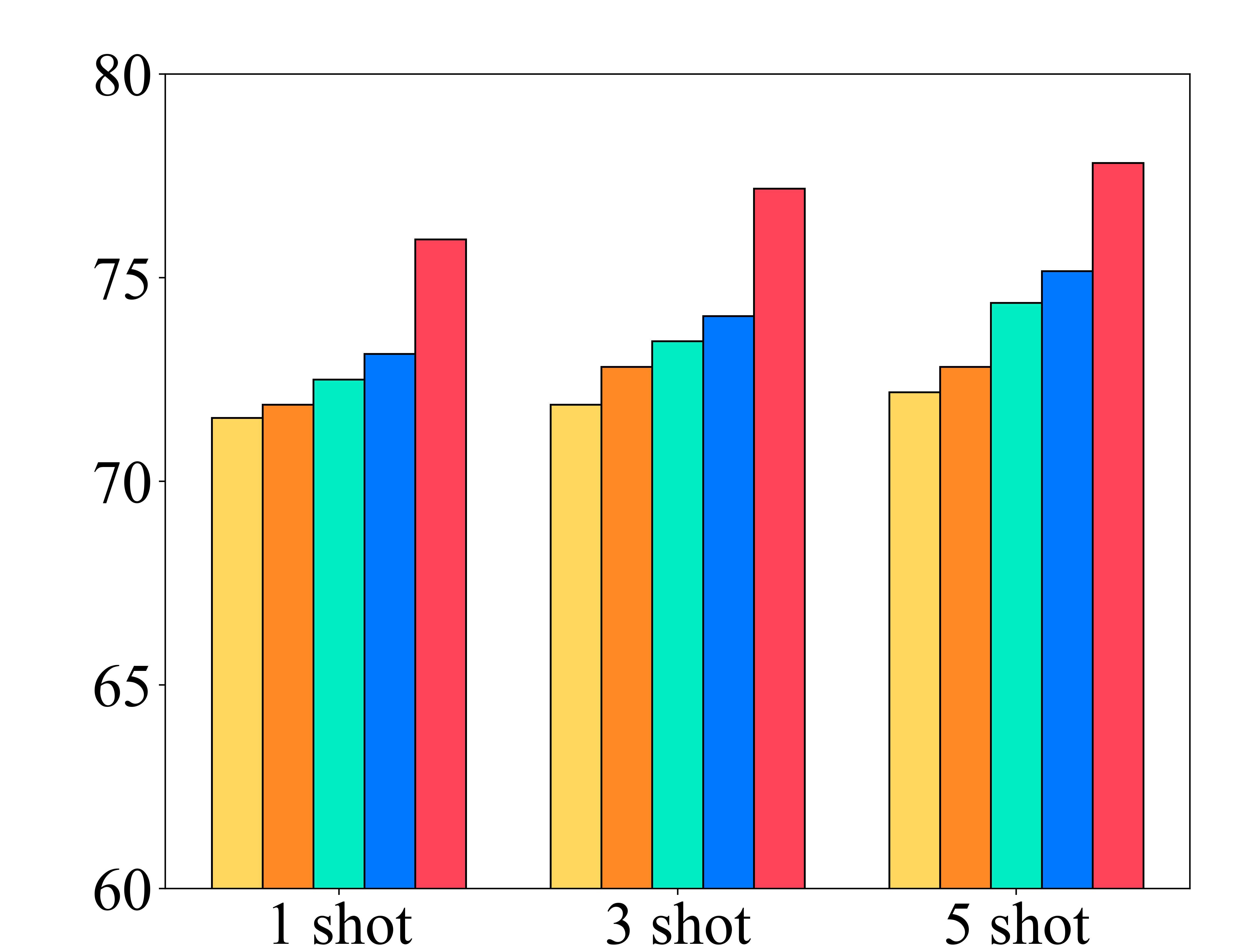}} \vspace{0.1cm} \\
	
	\caption{Visualization of the experiment results on unseen working conditions. The proposed REFML method outperforms other methods and provides an increase of accuracy by 2.17\% - 6.50\% compared to the FedProx-FT method. (a) Test on the CWRU dataset. (b) Test on the JNU dataset. (c) Test on the PHM2009 dataset. }
	\vspace{0.2cm}
	\label{fig6}
\end{figure*}

	\subsection{Visual Analyses}
	To provide an intuitive comparison of the feature extraction ability of different methods, the extracted features from different methods including raw data, FedAvg-FT, FedProx-FT, and the proposed REFML are illustrated in Fig. \ref{fig4} using the t-SNE technique, where different colors represent different health states. The feature information is contained in the output of the first FC layer in the models, which is a 256-dimensional vector, and the result of raw data is generated from a 1024-dimensional vector as its original length. \par
	It can be seen that features of the same health state learned by REFML are clustered well while features of different health states are separated well. In comparison, features learned by the other methods do not cluster well. For example, as shown in Fig. 4(c), the points of health state ``2" overlapped the points of the health state ``5", which implies that an effective classification cannot be achieved. 
	
	\subsection{Experiment Results on Unseen Working Conditions}
	In this part, we evaluate the proposed REFML method on unseen working conditions. 
	In each dataset, the working conditions of the meta-train and meta-test phases do not overlap.
	Among all the data corresponding to different working conditions, a portion of the data associated with certain working conditions is selected as data for the training clients, while another portion is chosen for the testing clients. Each client corresponds to a specific working condition. The meta-knowledge extraction is conducted in tasks of the training clients, and the generalization ability of the trained model is evaluated in the tasks of testing clients.
	Here we adopt $K$-fold cross-validation to conduct the experiment. For example, there are 4 working conditions in the CWRU dataset, and we have one client for each working condition. We choose clients 1, 2, and 3 as training clients and client 0 as testing client in the first fold, then we choose clients 2, 3, and 0 as training clients and client 1 as testing client in the second fold, and complete the experiment according to this rule. The performance of the model is then evaluated by averaging the results obtained from each fold. The specific experiment settings are listed in Table \ref{tab3}. 
	To clearly demonstrate this process, the diagnostic accuracy of the 1-shot sub-experiments conducted on the CWRU dataset are shown in Fig. \ref{fig5}. As we can see, although the generalization difficulty varies on different validation folds, for example, the performance of all algorithms is worse on the first fold than on other folds, the proposed REFML method achieves the best performance on all validation folds.\par
	
	The comparison results under different shot numbers with different methods are shown in Table \ref{tab4} and a visualization form in Fig. \ref{fig6}. It can be seen that the FedAvg method achieves the lowest accuracy while the FedProx method performs better because of its improvement against heterogeneity by restricting the local updates to be closer to the global model. Another observation is that the fine-tuned versions of FedAvg and FedProx outperform their original versions since they utilize the data on the testing task to adapt the model. Most importantly, the proposed REFML method achieves the best performance across all datasets and all shot numbers, and within a single dataset, the accuracy increases as the shot number increases. It provides an increase of accuracy by 2.17\%-6.50\% compared to the FedProx-FT method when generalizing to unseen working conditions under different shot numbers. In addition, as the shot number increases, the degree of performance improvement diminishes. For example, in the CWRU dataset, the performance increases from 1-shot to 3-shot and from 3-shot to 5-shot are 3.37\% and 1.30\%, respectively. Similarly, in the other two datasets, the corresponding performance gains drop from 3.89\% to 0.28\% and from 1.25\% to 0.63\%, respectively.\par
	
	\subsection{Experiment Results on Unseen Equipment Types}
	To further examine the effectiveness of the proposed method on generalizing to unseen equipment types, we conduct experiments across the CWRU dataset and the JNU dataset, where training data and testing data are collected from different types of bearings. To align classification categories, we selected four out of the ten health states from the CWRU dataset: three fault states with size 21 and an NA state, and kept all four health states in the JNU dataset. Thus, the tasks on these two datasets are all 4-way classification problems. The details of the designed experiments are listed in TABLE \uppercase\expandafter{\romannumeral5}.\par
	
	\setcounter{table}{4}

\begin{table}[t]
	\centering
	\caption{Experiment Settings on Unseen Equipment Types.}
	\label{tab5}
	\setlength{\tabcolsep}{5pt}
	\begin{tabular}{ ccc }
		\toprule
		Datasets & Meta-train conditions & Meta-test conditions \\
		\midrule
		CWRU to JNU & 0, 1, 2, 3  & 0, 1, 2  \\
		JNU to CWRU & 0, 1, 2  & 0, 1, 2, 3  \\
		\bottomrule
	\end{tabular}
\end{table}

\setcounter{table}{5}
\begin{table}[t]
	\centering
	\caption{Experiment Results on Unseen Equipment Types.}
	\label{tab6}
	\setlength{\tabcolsep}{5pt}
	\begin{tabular}{ ccccccc }
		
		\toprule
		\multirow{2}{*}{Methods} & \multicolumn{3}{c}{CWRU to JNU} & \multicolumn{3}{c}{JNU to CWRU}   \\
		& 1-shot & 3-shot & 5-shot & 1-shot & 3-shot & 5-shot \\
		
		\midrule
		FedAvg                                      & 44.99 & 45.33 & 45.67 & 49.85 & 50.16 & 50.47   \\
		FedAvg-FT                                & 50.83 & 51.67 & 54.17 & 60.32 & 66.10 & 67.50   \\
		FedProx                                     & 45.33 & 45.83 & 46.67 & 52.66 & 55.63 & 56.46   \\
		FedProx-FT                               & 51.67 & 52.92 & 53.33 & 63.91 & 72.19 & 72.30  \\
		REFML                                      & \textbf{68.06} & \textbf{68.89} & \textbf{69.17} & \textbf{78.55} & \textbf{85.63} & \textbf{90.63}   \\
		
		\bottomrule
		
	\end{tabular}
\end{table}

	\begin{figure}[h]
	\centering
    \vspace{-0.2cm}
	\subfloat[]{\includegraphics[width=0.7\columnwidth]{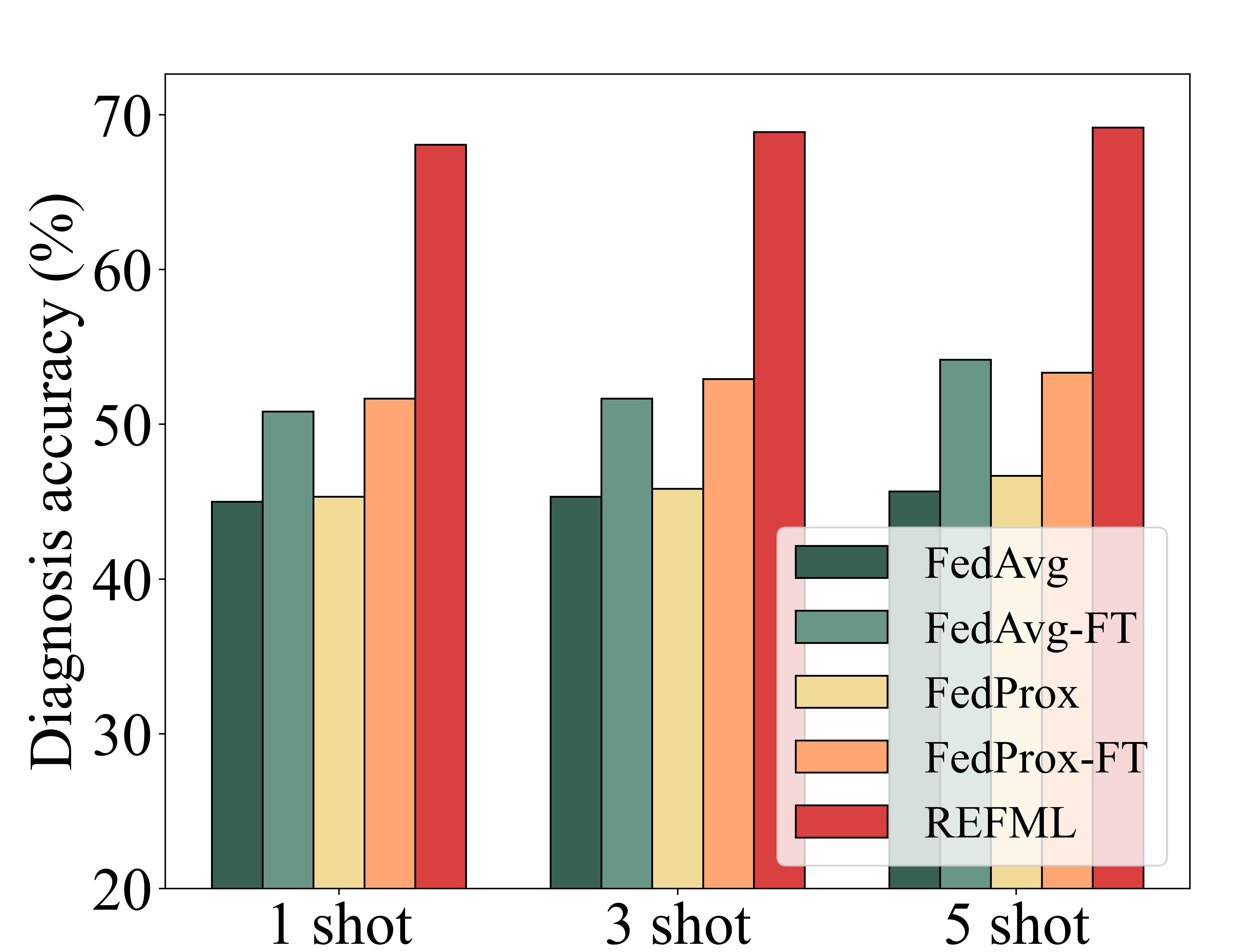}} \vspace{-0.2cm}\\
	\subfloat[]{\includegraphics[width=0.7\columnwidth]{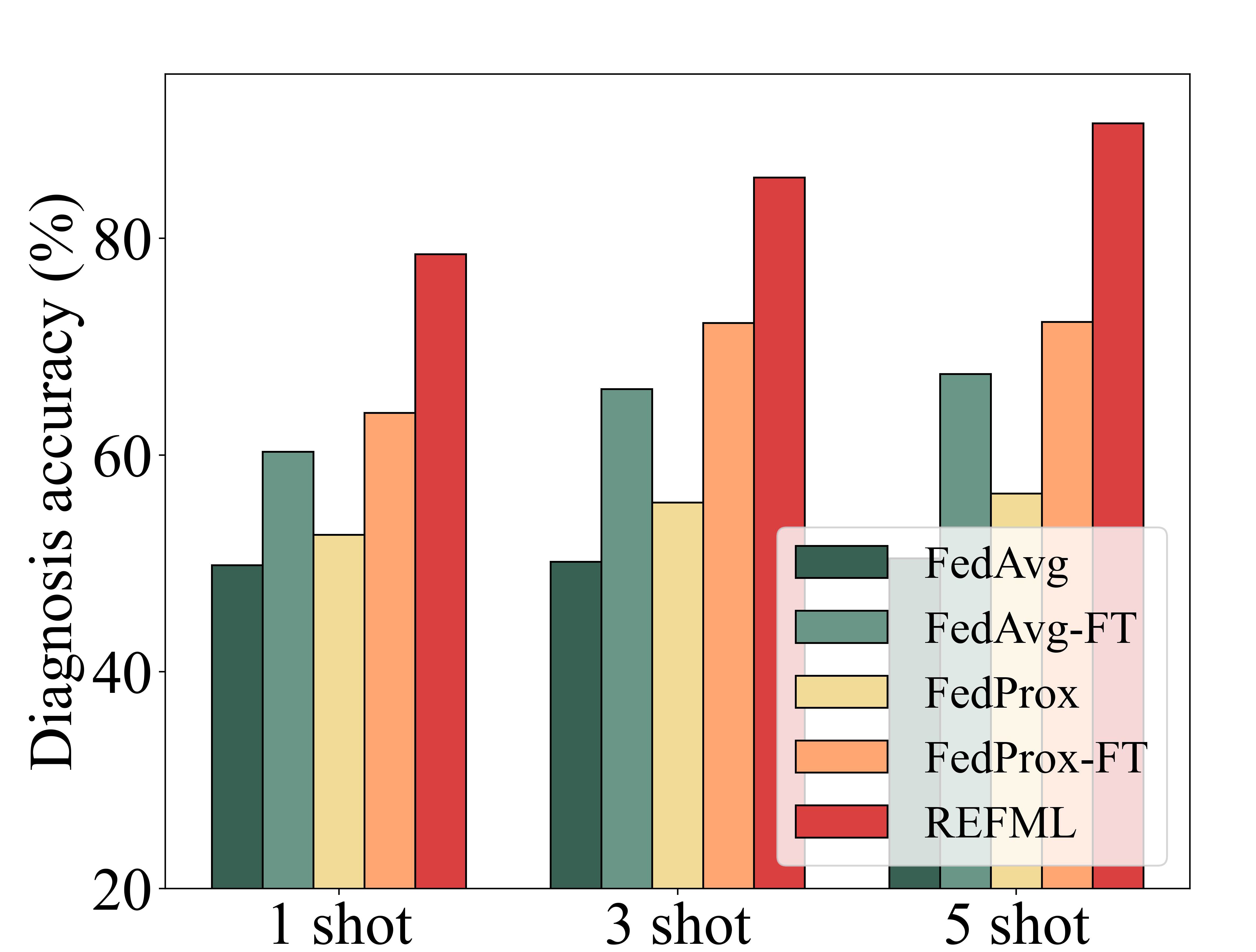}} \vspace{-0cm}\\
	
	\caption{Visualization of the experiment results on unseen equipment types. The proposed REFML method outperforms other methods and provides an increase of accuracy by 13.44\% - 18.33\% compared to the FedProx-FT method. (a) Generalization from CWRU dataset to JNU dataset. (b) Generalization from JNU dataset to CWRU dataset.}
	\label{fig7}
\end{figure}

	The comparison results under different shot numbers with different methods are shown in Table \uppercase\expandafter{\romannumeral6}, and Fig. \ref{fig7} presents a visualization of the results. It can be observed that the proposed REFML method still achieves the highest diagnostic accuracy and exhibits a positive correlation between its performance and the shot number. Another observation is the degree of performance improvement diminishes as the shot number becomes larger, which is consistent with the findings of the previous experiment. Furthermore, the performances of FedAvg-FT and FedProx-FT are significantly superior to those of their original versions, which indicates that the fine-tuning process has a considerable impact on accuracy improvement, suggesting the severity of domain discrepancy between training and testing. The advanced generalization ability of the REFML method is proved by 13.44\%-18.33\% accuracy improvement compared with the FedProx-FT method across 1-shot to 5-shot scenarios.\par

	\subsection{Ablation Analyses}
\begin{figure}[t]
	\vspace{0.25cm}
	\centering
	\includegraphics[width=0.7\columnwidth]{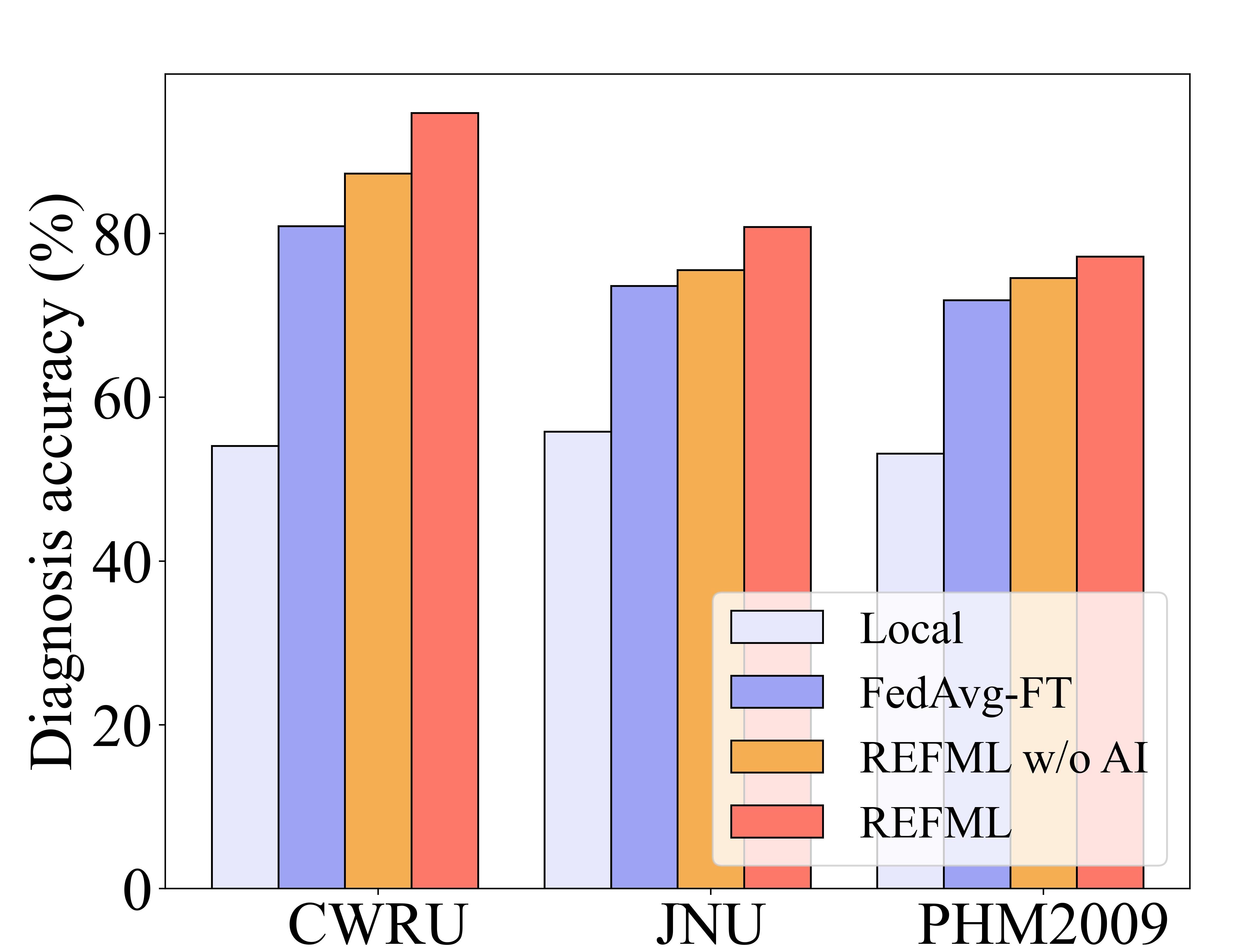}
	\caption{Results of the ablation experiments under 3-shot settings on three datasets.}
	\label{fig8}
	\vspace{0.2cm}
\end{figure}

	To verify the necessity of adopting the FL framework under privacy and security conditions and confirm each module proposed by us contributes to performance, including the training strategy based on meta-learning and representation encoding, and the adaptive interpolation module, the ablation experiments were designed. As depicted in Fig. \ref{fig8}, ``Local'' refers to testing clients performing local training without any federated communications. We also introduced a variant of the proposed REFML method called ``REFML w/o AI (adaptive interpolation) ", which omits the adaptive interpolation module before local training. \par
	
	We observe that the diagnostic accuracy of ``Local" is relatively low due to its inability to utilize information from the training clients, which reflects the effectiveness of FL. By employing our training strategy, the diagnostic accuracy of ``REFML w/o AI" surpasses FedAvg-FT across all datasets. This result indicates that our method can effectively extract domain-invariant representations for previously unseen classification tasks to achieve an accurate diagnosis. Moreover, the performance improvement of REFML compared with ``REFML w/o AI" demonstrates that our adaptive interpolation module can efficiently utilize local information, thereby mitigating the effects of domain discrepancy.\par
	
	The experimental results on both bearing data and gearbox data have already confirmed the effectiveness and robustness of the proposed method in the field of industrial FD, showing its potential for extensive applicability in industrial scenarios. It enables multiple clients to obtain a globally optimized model with high generalization performance. When encountering new working conditions and equipment types, it only requires a small number of samples to achieve good performance. \par
		
	Apart from industrial FD, our proposed approach can be readily applied to various applications with the same challenges of data privacy, domain discrepancy, and data scarcity. For instance, in the healthcare field, especially in image classification tasks, domain discrepancy occurs due to differences in imaging equipment and protocols across institutions.\par
	
	\section{Conclusion}
	In this paper, we have developed REFML to address the challenge of poor diagnosis performance in industrial FD caused by data privacy, domain discrepancy, and data scarcity. It enables multiple clients to collaboratively train a global model with high generalization ability while ensuring data privacy. The trained model can achieve superior performance on unobserved working conditions or equipment types even with limited training data. Specifically, a novel training strategy based on representation encoding and meta-learning has been invented, which harnesses data heterogeneity among training clients to facilitate OOD generalization in new clients by extracting meta-knowledge from different local diagnosis tasks and training a domain-invariant feature extractor. Furthermore, an adaptive interpolation method has been designed to calculate the optimal combination of local and global models before local training, which further utilizes local information to achieve better model performance.\par
	
	Experiments conducted on three real datasets have demonstrated that the proposed REFML method could perform well on unseen tasks using very limited training data with privacy guaranteed. Compared with the state-of-the-art methods, the proposed REFML framework achieves an increase in accuracy by 2.17\%-6.50\% when tested on unseen working conditions of the same equipment type and 13.44\%-18.33\% when tested on totally unseen equipment types, respectively.\par
		
	While our results demonstrate the promise of REFML, it suffers from high computational complexity due to the calculation of second-order derivatives in MAML. Future research includes exploring computationally efficient variants of MAML to enhance the practicality, scalability, and accuracy considering system heterogeneity.\par
	
		\bibliographystyle{IEEEtran}
\bibliography{myreference}

\end{document}